\documentclass[11pt]{article}

\usepackage[final]{acl}

\usepackage{times}
\usepackage{latexsym}

\usepackage[T1]{fontenc}

\usepackage[utf8]{inputenc}

\usepackage{microtype}

\usepackage{inconsolata}

\usepackage{graphicx}

\usepackage{booktabs}      
\usepackage{adjustbox}     
\usepackage{multirow}      
\usepackage{xcolor}        
\usepackage{colortbl}      
\usepackage{pgfplots}      
\pgfplotsset{compat=1.17}  
\usepackage{tcolorbox}

\definecolor{robust90}{RGB}{34,139,34}     
\definecolor{robust85}{RGB}{50,205,50}     
\definecolor{robust80}{RGB}{144,238,144}   
\definecolor{moderate75}{RGB}{255,255,150} 
\definecolor{moderate70}{RGB}{255,223,100} 
\definecolor{weak65}{RGB}{255,180,100}     
\definecolor{weak60}{RGB}{255,140,80}      
\definecolor{poor}{RGB}{255,100,100}       

\title{Evaluating Robustness of Large Language Models in Enterprise Applications: Benchmarks for Perturbation Consistency Across Formats and Languages}

\author{
 \textbf{Tara Bogavelli},
 \textbf{Oluwanifemi Bamgbose},
 \textbf{Gabrielle Gauthier Melançon},
\\
 \textbf{Fanny Riols},
 \textbf{Roshnee Sharma}
\\
 ServiceNow
\\
 \small{
   \textbf{Correspondence:} \href{tara.bogavelli@servicenow.com}{tara.bogavelli@servicenow.com}
 }
}

\begin{document}
\maketitle
\begin{abstract}
Enterprise LLM applications require consistently high quality and reliable performance across diverse scenarios, demanding robustness to minor variations. Existing research shows that even small prompt changes can lead to substantial differences in output, but has mainly focused on a narrow set of perturbations with small academic datasets, limiting their relevance to real-world applications. To address this, we present a comprehensive benchmark suite that evaluates robustness across multiple perturbation types, including general text edits (e.g., punctuation, whitespace), formatting changes (e.g., JSON, YAML), multilingual and cross-lingual inputs, and positional variations in instructions. Evaluating 11 models ranging from 4B to 120B+ parameters, we find that minor perturbations reduce performance by up to 40 percentage points on key enterprise metrics. Critically, we demonstrate that the relationship between model size and robustness is more nuanced than conventional assumptions suggest: an 8B parameter model (Ministral 3 8B) outperforms most larger models, while another 8B model (Llama 3.1 8B) performs worst overall. 
\end{abstract}

\section{Introduction}
\begin{figure}[t]
  \centering
  \includegraphics[width=\linewidth,trim={0.6cm 0 0.6cm 0},clip]{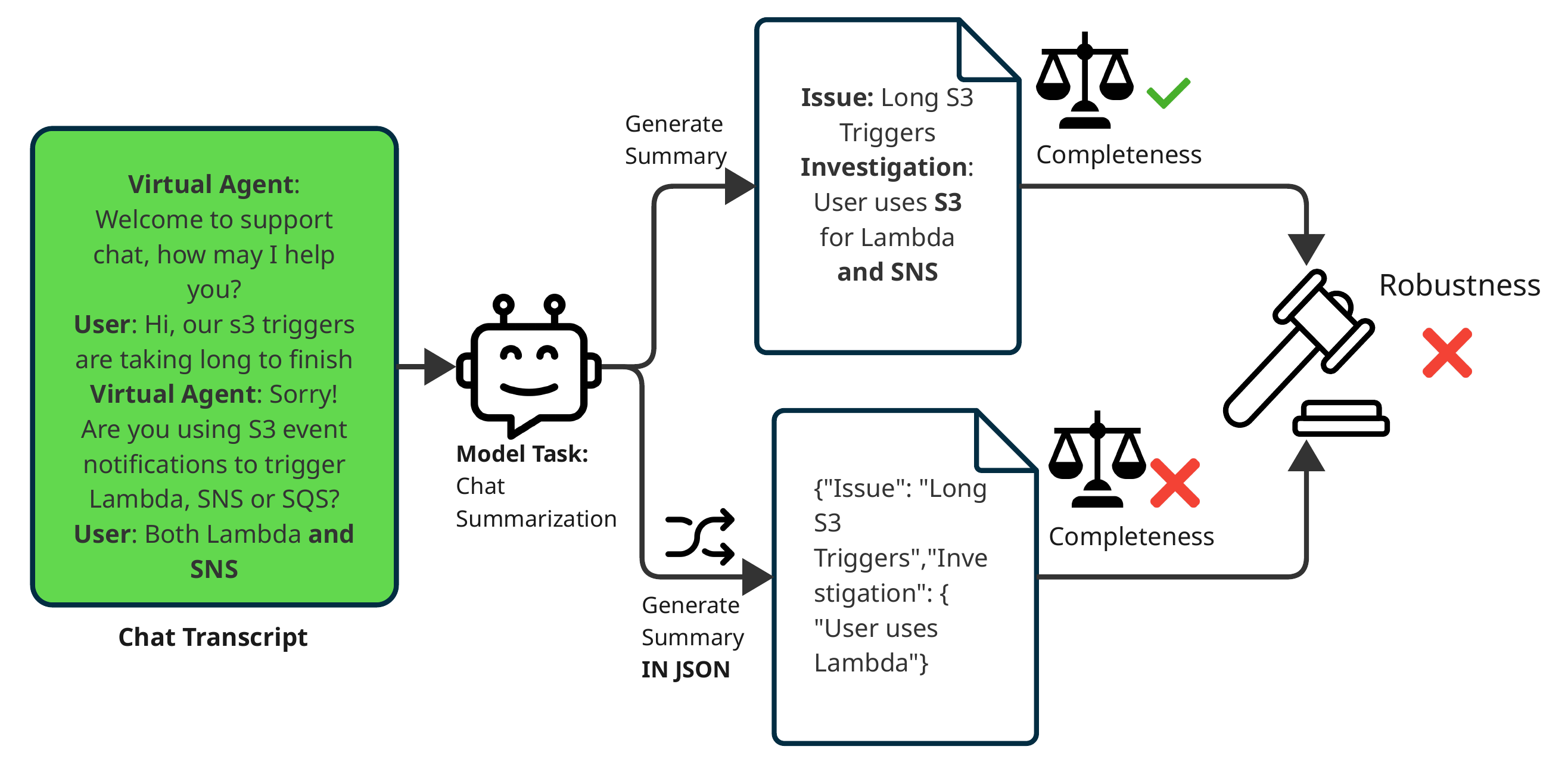}
  \caption{\textbf{JSON Format Perturbation on a Chat Summarization Task} The model performs a chat summarization task under two prompt conditions: standard (natural language) and perturbed (structured JSON format). We evaluate the outputs by measuring the quality (e.g., completeness) of each summary and assess robustness based on the consistency of model behavior across the two prompts.}
  \label{fig:model_teaser}
  \vspace{-3mm}
\end{figure}

Large language models (LLMs) are increasingly being used in enterprise environments to support high-stakes customer-facing applications such as virtual assistants, support automation, and content generation. However, users vary widely in their writing and understanding of model capabilities. In addition, prompt engineering practices often lack standardized guidelines or evaluation frameworks. As a result, we lack a clear understanding of how minor differences in prompt wording, structure, formatting or language affect model behavior and output quality. This variability introduces risks in consistency and performance across production use cases.
Prior work \citep{agrawal2025enhancingllmrobustnessperturbed, chaudhary2024understandingrobustnessllmbasedevaluations, singh2024robustnessllmsperturbationstext, wang2023largelanguagemodelsreally} has largely focused on sentence level perturbations (e.g. word swaps or paraphrasing) typically evaluated in narrow academic benchmarks. These studies do not adequately address the performance impact of prompt variation in large-scale, real-world enterprise contexts and on key metrics.
We investigate how minimal variations in prompts, such as changes in word order, formatting, punctuation, or language, affect the robustness of LLMs. Here, robustness is defined as the model’s ability to consistently generate outputs of similar quality when slight input modifications are applied.

To evaluate this, we first construct realistic enterprise datasets that reflect the complexity and noise typically associated with enterprise tasks—characteristics often underrepresented in existing academic benchmarks. We then establish a baseline by repeatedly running inference on identical prompts to measure consistency. Next, we introduce independent common enterprise variations—such as rewording, translation into other languages, tone shifts, and changes in output format (e.g., YAML versus plain text)—one at a time. As seen in Figure~\ref{fig:model_teaser}, for each variation, we compare the new response with the baseline to identify any differences, and then calculate a final robustness score based on both content changes and shifts in relevant use-case metrics (e.g Completeness, Faithfulness). Each evaluation is run separately to isolate the effect of each factor on model behavior and performance. 

Our central hypothesis is that in order to provide value in enterprise tasks, a model should maintain output consistency, task alignment and output quality when faced with non-substantive variations in prompt structure or language. Achieving such robustness is critical for scalable and dependable LLM deployment across teams, languages, and real-world applications. 

We observe that models are least robust to multilingual and cross-lingual perturbations (mean scores of 79.18 and 82.06 respectively), followed by positional changes (76.33). General perturbations(variation in whitespace, punctuation, etc)prove most manageable (85.51). Critically, we find that the relationship between model size and robustness is non-linear: while within-family scaling consistently improves robustness (GPT 5 nano to full: +6.66 points), training methodology can override size entirely—Ministral 3 8B outperforms GPT OSS 120B by 2.11 points, while another 8B model (Llama 3.1 8B) performs worst overall. This finding has significant practical implications for enterprise model selection. When using fixed prompt templates, models robust to positional changes are preferable. Tasks requiring specific file formats benefit from models resilient to format perturbations. Multilingual applications demand models with strong multilingual robustness—and, ideally, language-specific strengths. By mapping task requirements to model capabilities, our analysis supports informed model selection tailored to real-world enterprise constraints. We hope these findings help enterprise system designers more effectively integrate language models into their workflows by offering actionable insights into model selection and system design, grounded in an understanding of each model’s strengths and limitations under common real-world perturbations.

\section{Related Work}

Recent studies have systematically examined the robustness of large language models (LLMs) to various input perturbations, with a particular focus on task instructions, prompts, and evaluation reliability.

\citet{agrawal2025enhancingllmrobustnessperturbed} and \citet{chaudhary2024understandingrobustnessllmbasedevaluations} show that even minor lexical changes-such as typos or substitutions-can significantly degrade human-aligned metrics like coherence, fluency, and naturalness, often leading LLMs to emphasize edge-case behaviors inappropriately. They propose mitigation strategies like denoising, perplexity smoothing, and alignment techniques. Other studies address more naturalistic or synthetic text variations. \citet{singh2024robustnessllmsperturbationstext} find that LLMs exhibit general robustness to mild semantic-preserving noise (e.g. OCR errors, keyboard typos, word splits, and synonym swaps), but performance drops as perturbations become more complex or compounded. Similarly, \citet{wang2023largelanguagemodelsreally} focus on common lexical variations in user inputs, including word swaps, misspellings, and synonym replacements, revealing that minor lexical changes can reduce accuracy.

In the domain of task-specific robustness, \citet{chen2024nlperturbatorstudyingrobustnesscode} introduce 18 perturbation types relevant to code generation, including stylistic (e.g. active/passive voice) and structural (e.g. in docstrings) changes, reporting up to 21.2\% performance degradation. \citet{wang2024rupbenchbenchmarkingreasoningperturbations} evaluate LLM robustness to 9 adversarial textual perturbations - covering lexical (e.g. homophones, typos, leetspeak), syntactic (e.g. clefts, compound variations) and semantic (e.g. red herrings, CheckList, StressTest) levels - on reasoning datasets. The results show that larger models are generally more resilient, but commonsense and arithmetic tasks are highly sensitive due to reliance on contextual knowledge and precise computation. \citet{yang2025structevalbenchmarkingllmscapabilities} observe that LLMs struggle to generate renderable structured formats (YAML, JSON, CSV etc). In particular, models struggle more with generating these formats than with converting from one to another.

Other work investigates prompt sensitivity at scale. \citet{zhu2024promptrobustevaluatingrobustnesslarge} show that even small prompt perturbations, ranging from character-level (typos) to semantic-level (paraphrasing), can cause large performance drops, often transferable across models. \citet{cao2024structevaldeepenbroadenlarge} introduce metrics for format adherence and structural correctness and highlights substantial limitations in LLMs’ ability to reliably generate and translate between formats across 44 tasks and 18 formats.

In addition to lexical and task specific robustness, existing work demonstrates the importance of prompt ordering \citep{chen2024sifobenchmarkinvestigatingsequential, zheng2023judgingllmasajudgemtbenchchatbot, tang2024middlepermutationselfconsistencyimproves}. \citet{zeng2025ordermattersinvestigateposition} find that LLM performance varies based on constraint ordering in prompts. Specifically, they find that LLMs generally perform better when constraints are ordered from hard to easy.

While these studies offer valuable insights, they are typically limited to isolated perturbation types or narrow academic domains that do not represent real-world enterprise scenarios. Our work introduces a unified and comprehensive benchmark that evaluates LLM robustness across a broader range of realistic enterprise related perturbation types—including multilingual, positional and output-format variations.

\section{Proposed Method}

To systematically evaluate the robustness of language models in practical enterprise settings, we introduce a benchmark that spans four representative enterprise tasks:

\textbf{Case Summarization} – Given an enterprise support case (e.g., an issue reported by a customer or support agent), the model generates a concise summary highlighting the main problem, key events, and resolution steps. \textbf{Chat Summarization} – The model summarizes a multi-turn conversation between a user and a support agent, capturing the user’s issue, the agent’s responses, and the overall outcome. \textbf{Question and Answer (QnA)} – The model answers a user question by retrieving and synthesizing information from a provided set of reference documents. \textbf{Entity Slot-filling} – The model extracts entities from a natural language query and populates a predefined schema by filling in the appropriate slots.

For each task (see task prompt templates in Appendix ~\ref{appendix:prompts}), we create realistic input datasets and prompt the model on both the original inputs and their perturbed counterparts across five categories of perturbations: \textit{General}, \textit{Positional}, \textit{Format}, \textit{Multilingual}, and \textit{Cross-Lingual}. 

We evaluate robustness by comparing model outputs on original vs. perturbed inputs using a combination of content similarity and task-specific quality metrics. This two-pronged evaluation allows us to capture not only whether the output changes meaningfully, but also whether those changes impact task performance.

\subsection{Dataset Creation}
\begin{table}[ht]
\centering
\begin{tabular}{|c|p{6.5cm}|}
\hline
\textbf{ID} & \textbf{Issue} \\
\hline
1 & AWS IAM policy error on assume-role chaining \\
2 & BitLocker recovery prompt after Group Policy update \\
3 & Zscaler proxy blocks SharePoint file sync \\
4 & CrowdStrike policy not assigned to new OU \\
\hline
\end{tabular}
\caption{Sample rows from the dataset used in benchmarking.}
\label{tab:issue_samples}
\end{table}
We synthetically construct a base dataset comprising 420 IT-related issues reflecting common enterprise software failures using GPT 4.1. Each issue served as a seed for generating inputs across multiple downstream enterprise tasks. The base issues were manually curated and covered a range of realistic scenarios, such as authentication failures, policy misconfigurations, and connectivity disruptions. We will open-source our datasets and code upon acceptance. Table~\ref{tab:issue_samples} presents a few representative examples.

For each use case, we synthetically generate task-specific inputs from each issue in the base set e.g. case details for case summarization (See Table~\ref{tab:case_details}),  chat transcripts  for chat summarization (See Table~\ref{tab:chat_transcript}), queries and reference documents for QnA (See Table~\ref{tab:qna_kb_trunc}) and query-entity pairs (with expected entity lists) for entity slot-filling also using GPT 4.1(See Table~\ref{tab:entity_schema_ids}).

\subsubsection{Perturbation Types}

We introduce 5 categories of perturbations designed to test different aspects of LLM robustness. 

\textbf{General Perturbations} simulate natural variation in user inputs through word- and sentence-level changes. Subtypes include: Whitespace, Spelling, Punctuation, Casing, Paraphrasing of entire sentences, and Tone modifications. 

\textbf{Positional Perturbations} alter the order of sections within the prompt template to test instruction order sensitivity. We define 4 subtypes: Context (task input such as case details is moved in the prompt), Output Instructions (reordering or relocating formatting instructions like JSON guidelines), Reordering (changing the order of documents or bullet list items), and Mixed, which includes multiple positional changes from the other subtypes. 

\textbf{Format Perturbations} involve modifying the final output format instructions and comparing it to a free response format. Formats include JSON, HTML, XML, Markdown and YAML.

\textbf{Multilingual Perturbations} assess robustness across different input languages. Prompts are translated from English into the target language, and the model is instructed to respond in that same language. Languages include Spanish, French, French Canadian, German, Dutch, Italian, Japanese and Brazilian Portuguese. 

\textbf{Cross-Lingual Perturbations} test generalization across mismatched input and output languages. Each task’s prompt remains in English, while the contextual input (e.g. case details) is translated into one of the target languages, and the model is asked to respond in another. This creates n x n combinations, where n is the number of languages (here, 9), covering all possible input-output language pairs. 

\textbf{Baselines}. As a baseline, the same model is called twice with the identical prompt to estimate natural variation in generation. This allows us to distinguish sensitivity to perturbations from inherent sampling variance.

Each of these perturbation types is utilized in one of two distinct application scenarios. \textbf{User-input perturbations}: These included variations in tone, spelling, and language. They were applied directly to task inputs such as user queries or case details i.e only parts of the input that an end-user would be able to change in a real-world scenario. For multilingual evaluation, the English dataset was translated into each target language for all tasks.
\textbf{Prompt-template perturbations}: These involved altering prompt structure or formatting (e.g., changing instruction order or visual layout). Such modifications were applied dynamically during inference within our evaluation framework as prompt template modifications.

These perturbations were chosen to reflect real-world challenges that enterprise applications face: General perturbations test resilience to the natural variation in human communication and input quality variation; Positional perturbations evaluate whether models maintain consistent performance when prompt engineering decisions change or when different teams structure prompts differently; Format perturbations assess the model's ability to reliably produce structured outputs required for downstream system integration; Multilingual perturbations ensure consistent performance across global customer bases; and Cross-lingual perturbations test the model's ability to handle mixed-language scenarios common in international business contexts where input documents may be in one language while outputs are required in another.

\section{Evaluation Metrics}

We assess robustness using a two-stage evaluation process that focuses on content similarity and task-specific response quality. Using this content similarity and response quality, we report an overall score along with the content and quality deltas that contribute to it.

\subsection{Content Similarity}
While several similarity metrics were explored (see Appendix ~\ref{appendix:similarity_metrics}), we adopt \textbf{LLM-as-a-Judge} \citep{NEURIPS2023_91f18a12} as the primary method. In this setup, a large language model is prompted to evaluate whether the perturbed response differs meaningfully from the original, focusing on high-level \textbf{semantic consistency}, such as the preservation of key entities and facts, rather than superficial lexical overlap. The scoring scale is defined as follows:

\begin{itemize}
    \item \textbf{3} - The responses are semantically identical; no meaningful difference in content.
    \item \textbf{2} - Minor but noticeable differences in meaning; some content may be altered or omitted.
    \item \textbf{1} - Major differences in meaning; the core message or factual content is changed.
\end{itemize}

(See 'Robustness Judge Prompt' in Appendix ~\ref{appendix:prompts}).

\subsection{Quality Metrics}
\label{appendix:quality_metrics}
If the similarity score is less than 3, indicating a meaningful shift in content, we evaluate both responses using task-specific \textbf{quality metrics}, and compute the delta to determine if their quality differs. These quality metrics represent how well the model has performed its use-case task based on specific criteria. Below are the quality metrics used for each task:

\begin{itemize}
    \item \textbf{Case Summarization}: Faithfulness, Completeness
    \item \textbf{Chat Summarization}: Faithfulness, Completeness
    \item \textbf{Q\&A}: Faithfulness, Completeness, Relevance, Citations, Conciseness
    \item \textbf{Entity Slot-filling}: Exact-match F1 score and ROUGE-L on extracted entities
\end{itemize}

\textit{Faithfulness} represents the extent to which the model's response adheres to the information provided in the input prompt and context, without including unsupported claims, factual errors, or hallucinations. 

\textit{Completeness} represents the degree to which the model response covers all important and useful content that was provide in the context in the final response. 

\textit{Relevance} evaluates how well the model’s response remains directly connected to the user’s query, avoiding repetition, redundancy, and the inclusion of unrelated information. 

\textit{Citations} assesses whether factual claims in the model’s response are properly supported by, and attributed to, the appropriate sources within the provided documents.

\textit{Conciseness} checks whether the model response is succinct, to the point, and avoids excessive rambling.

\textit{Entity Match} is defined as the F1 score between entities in the candidate and response, where entities are key-value pairs extracted from user input according to a schema. Keys must match exactly. For values, we require exact matches if the value has predefined choices in the schema, and use a ROUGE-L F1 threshold of 0.7 for free-text (description) fields.

These quality metrics were selected based on task-specific indicators of successful LLM behavior that could correlate with user satisfaction. For example, proper citation is important in question and answering tasks where a customer may want to refer to certain documents based on the model's response for additional information. This would not be as relevant in another task like summarization. We find that completeness and faithfulness are important for all tasks as model responses with hallucinations are always considered problematic and incomplete responses are generally unhelpful to the user. For a task like entity slot-filling where there is exactly one correct answer, we use exact match instead of LLM as a Judge for reliability. 

All of the above metrics (except entity match) are represented as Judge Prompts within an LLM-as-a-Judge setting (See prompts in ~\ref{appendix:prompts}). We prompt the Judge Model to rate each model response on a scale of 1-3 for the given metric and use GPT 4.1 as the Judge Model.

\subsection{Overall Score}
To quantify the effects of perturbations, we compute content delta, quality delta and an overall robustness score derived from the content similarity and task-specific quality scores:

\textbf{Content Delta} measures how often a perturbation causes a meaningful change in response content. Given a dataset of $N$ samples, we apply a specific perturbation (e.g., case change or synonym replacement) to generate $N$ perturbed versions. For each original-perturbed pair, the LLM judge assigns a similarity score (1–3). Content Delta is calculated as the proportion of samples with a score \textless{} 3. 

\textit{Example}:  
Consider 100 original-perturbed pairs. If 30 receive a similarity score of 2 and 20 receive a score of 1, then 50 out of 100 pairs show meaningful content differences. Thus, the Content Delta is:
\[
\Delta_{\textit{Content}} = \frac{30 + 20}{100} = 0.5
\]

\textbf{Quality Delta} captures the proportion of content-shifted pairs where the perturbation also affects task-specific output quality. We focus only on the subset of pairs with a similarity score \textless{} 3. For each such pair, we evaluate the original and perturbed responses using the appropriate quality metrics. If any metric changes—regardless of direction—we count it as a quality change. 

\textit{Example}:  
Continuing the example above, suppose that among the 50 content-shifted pairs, 15 show a change in at least one quality metric (e.g., a drop in faithfulness or a gain in relevance). Then the Quality Delta is:
\[
\Delta_{\textit{Quality}} = \frac{15}{50} = 0.3
\]

Finally, we compute the \textbf{overall robustness score} using the following formula:

\begin{equation}
\textit{Overall Score} = 1 - (\Delta_{\textit{Content}} \times \Delta_{\textit{Quality}})
\label{eq:robustness_score}
\end{equation}

This formulation penalizes both frequent and impactful content changes. A higher score indicates greater robustness, reflecting both the likelihood and severity of performance degradation under input perturbations.

\section{Experiments}

\subsection{Models}
The benchmark was evaluated on 11 models spanning a range of sizes (4B to 120B+ parameters) and architectures. We include both open-source and proprietary models: GPT 5 family (Nano, Mini, 5.2), GPT 4.1, GPT OSS (20B, 120B), Gemma 3 (4B, 12B), Ministral 3 8B Instruct 2512, Mistral Small 3.2 24B Instruct 2506, and Llama 3.1 8B Instruct.

For each model, parameters were chosen to ensure sufficient token capacity, reproducibility, and coherent responses. A minimum token limit of 6,000 was used for non-reasoning models and 12,000 for reasoning models. All models were run with a temperature of 0 and a fixed seed (where these parameters were supported) to maximize determinism. All reasoning models were run on with the lowest reasoning setting available. All results report mean and standard deviation across 5 independent runs.

\subsection{Overall Results Analysis}

\definecolor{robust90}{RGB}{34,139,34}    
\definecolor{robust85}{RGB}{50,205,50}    
\definecolor{robust80}{RGB}{144,238,144}  
\definecolor{moderate75}{RGB}{255,255,150} 
\definecolor{moderate70}{RGB}{255,223,100} 
\definecolor{weak65}{RGB}{255,180,100}    
\definecolor{weak60}{RGB}{255,140,80}     
\definecolor{poor}{RGB}{255,100,100}      

\begin{table*}[t]
\centering
\small
\begin{adjustbox}{max width=\textwidth}
\begin{tabular}{@{}llccccccccc@{}}
\toprule
\textbf{Model} & \textbf{Size} & \textbf{Overall} & \textbf{Content} & \textbf{Quality} & \textbf{Baseline} & \textbf{General} & \textbf{Positional} & \textbf{Format} & \textbf{Multilingual} & \textbf{Crosslingual} \\
 & & \textbf{Score} $\uparrow$ & \textbf{Delta} $\downarrow$ & \textbf{Delta} $\downarrow$ & & & & & & \\
\midrule
GPT 5.2 & Large 
  & \cellcolor{robust90!70}\textbf{91.01}{\scriptsize±0.07} 
  & \textbf{14.34}{\scriptsize±0.09} 
  & 63.07{\scriptsize±0.69} 
  & \cellcolor{robust90!70}93.27{\scriptsize±0.48} 
  & \cellcolor{robust90!70}90.68{\scriptsize±0.15} 
  & \cellcolor{robust90!70}\textbf{91.22}{\scriptsize±0.17} 
  & \cellcolor{robust90!70}\textbf{91.25}{\scriptsize±0.43} 
  & \cellcolor{robust85!70}\textbf{89.06}{\scriptsize±0.14} 
  & \cellcolor{robust90!70}90.56{\scriptsize±0.09} \\

GPT 4.1 & Large 
  & \cellcolor{robust90!70}90.20{\scriptsize±0.60} 
  & 23.33{\scriptsize±1.91} 
  & 45.78{\scriptsize±2.49} 
  & \cellcolor{robust90!70}95.31{\scriptsize±0.65} 
  & \cellcolor{robust90!70}90.07{\scriptsize±0.24} 
  & \cellcolor{robust85!70}86.89{\scriptsize±0.62} 
  & \cellcolor{robust90!70}91.00{\scriptsize±0.34} 
  & \cellcolor{robust85!70}86.82{\scriptsize±0.59} 
  & \cellcolor{robust90!70}\textbf{91.36}{\scriptsize±1.58} \\

Ministral 3 8B & 8B 
  & \cellcolor{robust85!70}89.40{\scriptsize±0.73} 
  & 29.37{\scriptsize±1.98} 
  & \textbf{34.23}{\scriptsize±2.51} 
  & \cellcolor{robust90!70}\textbf{98.32}{\scriptsize±0.48} 
  & \cellcolor{robust90!70}\textbf{92.49}{\scriptsize±0.25} 
  & \cellcolor{moderate75!70}78.92{\scriptsize±0.34} 
  & \cellcolor{robust85!70}89.45{\scriptsize±0.31} 
  & \cellcolor{robust85!70}89.12{\scriptsize±0.04} 
  & \cellcolor{robust85!70}87.66{\scriptsize±4.61} \\

GPT 5 Mini & Med 
  & \cellcolor{robust85!70}88.55{\scriptsize±0.15} 
  & 19.36{\scriptsize±0.06} 
  & 59.90{\scriptsize±0.88} 
  & \cellcolor{robust90!70}93.86{\scriptsize±0.70} 
  & \cellcolor{robust85!70}87.96{\scriptsize±0.50} 
  & \cellcolor{robust85!70}87.87{\scriptsize±0.22} 
  & \cellcolor{robust85!70}86.36{\scriptsize±0.29} 
  & \cellcolor{robust85!70}85.74{\scriptsize±0.17} 
  & \cellcolor{robust85!70}89.53{\scriptsize±0.10} \\

GPT OSS 120B & 120B 
  & \cellcolor{robust85!70}87.29{\scriptsize±2.54} 
  & 28.79{\scriptsize±2.24} 
  & 43.78{\scriptsize±3.28} 
  & \cellcolor{robust90!70}93.54{\scriptsize±0.36} 
  & \cellcolor{robust85!70}88.21{\scriptsize±1.87} 
  & \cellcolor{moderate70!70}73.00{\scriptsize±20.91} 
  & \cellcolor{robust85!70}89.10{\scriptsize±0.71} 
  & \cellcolor{robust85!70}85.78{\scriptsize±0.31} 
  & \cellcolor{robust85!70}88.53{\scriptsize±1.17} \\

GPT 5 Nano & Small 
  & \cellcolor{robust80!70}84.35{\scriptsize±0.14} 
  & 31.61{\scriptsize±0.31} 
  & 50.23{\scriptsize±0.80} 
  & \cellcolor{robust90!70}92.95{\scriptsize±0.60} 
  & \cellcolor{robust85!70}85.31{\scriptsize±0.30} 
  & \cellcolor{robust80!70}83.04{\scriptsize±0.23} 
  & \cellcolor{robust80!70}82.06{\scriptsize±0.29} 
  & \cellcolor{moderate75!70}77.67{\scriptsize±0.30} 
  & \cellcolor{robust85!70}85.04{\scriptsize±0.24} \\

Mistral Small 3.2 24B & 24B 
  & \cellcolor{robust80!70}82.45{\scriptsize±0.12} 
  & 41.33{\scriptsize±0.19} 
  & 41.37{\scriptsize±0.57} 
  & \cellcolor{robust90!70}96.58{\scriptsize±0.26} 
  & \cellcolor{robust80!70}84.50{\scriptsize±0.23} 
  & \cellcolor{moderate70!70}74.31{\scriptsize±0.26} 
  & \cellcolor{robust85!70}85.26{\scriptsize±0.13} 
  & \cellcolor{moderate75!70}75.22{\scriptsize±0.20} 
  & \cellcolor{moderate75!70}78.82{\scriptsize±0.06} \\

GPT OSS 20B & 20B 
  & \cellcolor{robust80!70}82.03{\scriptsize±5.34} 
  & 42.28{\scriptsize±4.26} 
  & 42.20{\scriptsize±6.36} 
  & \cellcolor{robust90!70}92.09{\scriptsize±2.01} 
  & \cellcolor{robust80!70}83.26{\scriptsize±6.51} 
  & \cellcolor{weak65!70}66.29{\scriptsize±18.93} 
  & \cellcolor{robust85!70}85.20{\scriptsize±6.50} 
  & \cellcolor{robust80!70}82.07{\scriptsize±0.17} 
  & \cellcolor{robust80!70}83.28{\scriptsize±2.89} \\

Gemma 3 12B & 12B 
  & \cellcolor{moderate75!70}79.48{\scriptsize±2.03} 
  & 41.27{\scriptsize±1.63} 
  & 48.24{\scriptsize±2.51} 
  & \cellcolor{robust90!70}95.32{\scriptsize±0.83} 
  & \cellcolor{robust80!70}82.88{\scriptsize±0.17} 
  & \cellcolor{weak65!70}68.53{\scriptsize±9.47} 
  & \cellcolor{moderate75!70}75.49{\scriptsize±6.94} 
  & \cellcolor{moderate75!70}75.25{\scriptsize±1.82} 
  & \cellcolor{moderate75!70}79.83{\scriptsize±1.69} \\

Gemma 3 4B & 4B 
  & \cellcolor{moderate75!70}75.92{\scriptsize±0.20} 
  & 51.08{\scriptsize±0.20} 
  & 45.90{\scriptsize±0.45} 
  & \cellcolor{robust90!70}93.45{\scriptsize±0.56} 
  & \cellcolor{moderate75!70}79.15{\scriptsize±0.27} 
  & \cellcolor{weak65!70}66.26{\scriptsize±0.16} 
  & \cellcolor{moderate70!70}74.78{\scriptsize±0.39} 
  & \cellcolor{weak65!70}67.24{\scriptsize±0.24} 
  & \cellcolor{moderate70!70}74.63{\scriptsize±0.16} \\

Llama 3.1 8B & 8B 
  & \cellcolor{weak65!70}69.52{\scriptsize±0.30} 
  & 64.36{\scriptsize±0.47} 
  & 45.45{\scriptsize±0.71} 
  & \cellcolor{robust90!70}92.80{\scriptsize±1.01} 
  & \cellcolor{moderate75!70}76.15{\scriptsize±0.43} 
  & \cellcolor{weak60!70}63.33{\scriptsize±0.52} 
  & \cellcolor{moderate70!70}74.49{\scriptsize±0.44} 
  & \cellcolor{poor!70}56.97{\scriptsize±0.40} 
  & \cellcolor{poor!70}53.40{\scriptsize±0.46} \\

\midrule
\textit{Average} & — & 83.65{\scriptsize±1.11} & 35.19{\scriptsize±1.21} & 47.28{\scriptsize±1.93} & 94.32{\scriptsize±0.72} & 85.51{\scriptsize±0.99} & 76.33{\scriptsize±4.71} & 84.04{\scriptsize±1.53} & 79.18{\scriptsize±0.40} & 82.06{\scriptsize±1.19} \\
\bottomrule
\end{tabular}
\end{adjustbox}
\caption{Overall Scores (mean ± std over 5 runs) per model across perturbation categories. Cell colors indicate robustness level for category scores: \colorbox{robust90!70}{$\geq$90}, \colorbox{robust85!70}{85--90}, \colorbox{robust80!70}{80--85}, \colorbox{moderate75!70}{75--80}, \colorbox{moderate70!70}{70--75}, \colorbox{weak65!70}{65--70}, \colorbox{weak60!70}{60--65}, \colorbox{poor!70}{$<$60}. Content and Quality Delta columns are uncolored as lower values are better. Models sorted by Overall Score; best per column in \textbf{bold}.}
\label{tab:main_results}
\end{table*}

\begin{table}[t]
\centering
\small
\begin{tabular}{@{}lccc@{}}
\toprule
\textbf{Category} & \textbf{Mean} & \textbf{Mean Drop} & \textbf{Mean Std} \\
 & \textbf{Score} & \textbf{from Base} & \textbf{(Dev)} \\
\midrule
Baseline & 94.32 & — & 0.72 \\
General & 85.51 & -8.81 & 0.99 \\
Format & 84.04 & -10.28 & 1.53 \\
Crosslingual & 82.06 & -12.26 & 1.19 \\
Positional & 76.33 & -17.99 & 4.71 \\
Multilingual & 79.18 & -15.14 & 0.40 \\
\bottomrule
\end{tabular}
\caption{Perturbation category difficulty analysis. Categories ordered by mean overall score drop from baseline. Higher standard deviation across models indicates less predictable difficulty. Multilingual and positional perturbations are universally challenging; general perturbations are manageable for most models.}
\label{tab:category_difficulty}
\end{table}

Table~\ref{tab:main_results} presents the main results across all models and perturbation categories, with cell coloring indicating robustness levels. Several key findings emerge:

\textbf{Top performers.} GPT 5.2 achieves the highest overall robustness score (91.01±0.07), followed by GPT 4.1 (90.20±0.60) and Ministral 3 8B (89.40±0.73). These models maintain consistent performance across perturbation types, with Content Deltas below 30\%. The heatmap reveals that top models show uniformly green (robust) coloring across categories.

\textbf{Bottom performers.} Llama 3.1 8B performs worst overall (69.52±0.30), with a Content Delta of 64.36\%—indicating that nearly two-thirds of perturbed inputs produce meaningfully different outputs. The heatmap highlights catastrophic failures (red cells) for this model on multilingual (56.97) and crosslingual (53.40) perturbations.

\textbf{Perturbation difficulty hierarchy.} Across all models, we observe a consistent difficulty ordering visible in the column-wise color patterns: baseline scores are uniformly high (green), while positional and multilingual columns show the most orange/yellow cells. Table~\ref{tab:category_difficulty} quantifies this: multilingual and positional perturbations cause the largest performance drops (-15.14 and -17.99 points respectively), followed by crosslingual (-12.26) and format (-10.28). General perturbations prove most manageable (-8.81 points).

\subsection{Model Size Analysis}

Conventional wisdom suggests that larger models exhibit greater robustness. Our results partially confirm but significantly complicate this view.

\textbf{Within-family scaling helps.} The GPT 5 family shows robustness improvements from scaling: nano (84.35±0.14) $\rightarrow$ mini (88.55±0.15) $\rightarrow$ 5.2 (91.01±0.07), an overall improvement of 6.66-point with consistently low variance.. Similarly, Gemma 3 improves from 4B (75.92±0.20) to 12B (79.48±2.03), and GPT OSS from 20B (82.03±5.34) to 120B (87.29±2.54).

\textbf{Architecture and training dominate scale.} The most striking finding is the 19.88-point gap between two 8B models: Ministral 3 8B (89.40±0.73) dramatically outperforms Llama 3.1 8B (69.52±0.30). Furthermore, Ministral 3 8B outperforms the 120B GPT OSS model by 2.11 points. This demonstrates that training methodology, data quality, and architecture choices can override substantial size advantages.

\textbf{Variance patterns.} The GPT OSS family exhibits notably high variance (±5.34 for 20B, ±2.54 for 120B), suggesting less stable behavior compared to the GPT 5 family (all $<$±0.15). This variance is particularly pronounced for positional perturbations (see Figure~\ref{fig:positional_scores}).

\subsection{Perturbation Category Analysis}

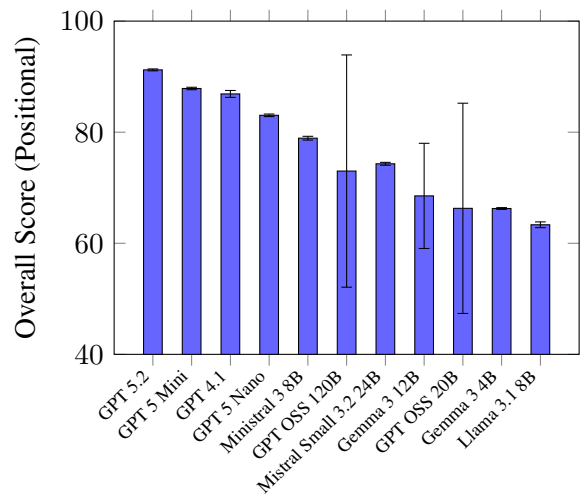
\begin{figure}[t]
\centering
\begin{tikzpicture}
\begin{axis}[
    ybar,
    width=\columnwidth,
    height=6cm,
    ylabel={Overall Score (Positional)},
    symbolic x coords={GPT 5.2, GPT 5 Mini, GPT 4.1, GPT 5 Nano, Ministral 3 8B, GPT OSS 120B, Mistral Small 3.2  24B, Gemma 3 12B, GPT OSS 20B, Gemma 3 4B, Llama 3.1 8B},
    xtick=data,
    x tick label style={rotate=45, anchor=east, font=\scriptsize},
    ymin=40, ymax=100,
    bar width=7pt,
    error bars/y dir=both,
    error bars/y explicit,
]
\addplot[fill=blue!60, error bars/.cd, y dir=both, y explicit] coordinates {
    (GPT 5.2, 91.22) +- (0, 0.17)
    (GPT 5 Mini, 87.87) +- (0, 0.22)
    (GPT 4.1, 86.89) +- (0, 0.62)
    (GPT 5 Nano, 83.04) +- (0, 0.23)
    (Ministral 3 8B, 78.92) +- (0, 0.34)
    (Mistral Small 3.2 24B, 74.31) +- (0, 0.26)
    (GPT OSS 120B, 73.00) +- (0, 20.91)
    (Gemma 3 12B, 68.53) +- (0, 9.47)
    (GPT OSS 20B, 66.29) +- (0, 18.93)
    (Gemma 3 4B, 66.26) +- (0, 0.16)
    (Llama 3.1 8B, 63.33) +- (0, 0.52)
};
\end{axis}
\end{tikzpicture}
\caption{Overall Scores on positional perturbations by model (mean ± std). The GPT 5 family demonstrates consistent advantages with low variance. Note the extremely high variance for GPT OSS 20B (±18.93) and GPT OSS 120B (±20.91), indicating unstable positional handling. Smaller open-source models (Llama 3.1 8B, Gemma 3 4B) show significant sensitivity to instruction ordering.}
\label{fig:positional_scores}
\end{figure}

\begin{table*}[t]
\centering
\small
\begin{adjustbox}{max width=\textwidth}
\begin{tabular}{@{}lcccccccc@{}}
\toprule
\textbf{Model} & \textbf{German} & \textbf{Spanish} & \textbf{French} & \textbf{Italian} & \textbf{Japanese} & \textbf{Dutch} & \textbf{Portuguese} & \textbf{Brazilian} \\
& & & & & & & & \textbf{Portuguese} \\
\midrule
GPT 5.2 & 63.34{\scriptsize±1.84} & 57.37{\scriptsize±3.24} & 56.79{\scriptsize±2.37} & 60.46{\scriptsize±1.54} & \textbf{64.77}{\scriptsize±2.88} & 63.75{\scriptsize±2.20} & 62.96{\scriptsize±2.41} & 61.93{\scriptsize±1.56} \\
GPT 4.1 & 37.45{\scriptsize±1.50} & 33.65{\scriptsize±0.58} & 33.47{\scriptsize±1.47} & \textbf{40.01}{\scriptsize±2.32} & 29.14{\scriptsize±2.60} & 39.29{\scriptsize±1.45} & 38.48{\scriptsize±2.80} & 35.44{\scriptsize±2.73} \\
GPT 5 Mini & 53.40{\scriptsize±1.06} & 54.28{\scriptsize±1.13} & 50.52{\scriptsize±1.24} & 56.75{\scriptsize±2.36} & 51.20{\scriptsize±0.95} & 55.66{\scriptsize±1.23} & \textbf{56.93}{\scriptsize±0.56} & 54.50{\scriptsize±1.06} \\
GPT 5 Nano & \textbf{54.45}{\scriptsize±1.64} & 47.22{\scriptsize±0.66} & 45.21{\scriptsize±1.94} & 48.76{\scriptsize±1.03} & 45.50{\scriptsize±1.73} & 53.54{\scriptsize±1.14} & 49.44{\scriptsize±2.26} & 47.48{\scriptsize±1.02} \\
Ministral 3 8B & 26.69{\scriptsize±1.28} & 27.64{\scriptsize±1.41} & 27.96{\scriptsize±0.54} & 27.54{\scriptsize±1.12} & \textbf{31.31}{\scriptsize±1.39} & 26.15{\scriptsize±1.41} & 29.48{\scriptsize±1.25} & 27.90{\scriptsize±1.58} \\
GPT OSS 120B & 39.01{\scriptsize±0.61} & 35.07{\scriptsize±1.05} & 33.82{\scriptsize±1.04} & 40.05{\scriptsize±1.90} & 40.35{\scriptsize±1.47} & \textbf{43.67}{\scriptsize±1.06} & 35.80{\scriptsize±1.65} & 36.06{\scriptsize±1.34} \\
Mistral Small 3.2 24B & 43.04{\scriptsize±1.37} & 36.67{\scriptsize±0.59} & 41.23{\scriptsize±0.47} & 38.69{\scriptsize±0.66} & \textbf{45.74}{\scriptsize±0.84} & 37.90{\scriptsize±0.81} & 37.41{\scriptsize±0.48} & 36.43{\scriptsize±1.27} \\
GPT OSS 20B & 33.17{\scriptsize±1.66} & 28.60{\scriptsize±0.95} & 28.50{\scriptsize±0.83} & 30.87{\scriptsize±1.07} & \textbf{33.88}{\scriptsize±0.42} & 31.23{\scriptsize±0.94} & 29.00{\scriptsize±1.18} & 29.39{\scriptsize±1.30} \\
Gemma 3 12B & 40.77{\scriptsize±4.17} & 43.26{\scriptsize±8.52} & 36.81{\scriptsize±0.66} & 40.64{\scriptsize±1.04} & \textbf{44.88}{\scriptsize±1.02} & 40.74{\scriptsize±0.90} & 39.67{\scriptsize±2.00} & 40.36{\scriptsize±2.78} \\
Gemma 3 4B & 44.81{\scriptsize±0.38} & \textbf{48.62}{\scriptsize±0.75} & 40.89{\scriptsize±0.37} & 45.32{\scriptsize±0.95} & 45.95{\scriptsize±0.36} & 47.51{\scriptsize±0.79} & 42.10{\scriptsize±0.67} & 43.36{\scriptsize±1.09} \\
Llama 3.1 8B & 46.40{\scriptsize±0.81} & 44.73{\scriptsize±1.24} & 44.90{\scriptsize±0.84} & 46.40{\scriptsize±0.75} & \textbf{54.75}{\scriptsize±0.95} & 47.99{\scriptsize±0.36} & 44.03{\scriptsize±1.28} & 45.88{\scriptsize±1.57} \\
\midrule
\textit{Mean} & 43.87 & 41.56 & 40.01 & 43.23 & \textbf{44.32} & 44.31 & 42.30 & 41.70 \\
\bottomrule
\end{tabular}
\end{adjustbox}
\caption{Multilingual Quality Delta by language (mean ± std; lower is better, indicates less quality change when perturbed). \textbf{Bold} in data rows indicates worst (highest) delta for that model.} 
\label{tab:multilingual}
\end{table*}

\begin{table*}[t]
\centering
\small
\begin{tabular}{@{}lccccc@{}}
\toprule
\textbf{Model} & \textbf{JSON} & \textbf{YAML} & \textbf{XML} & \textbf{Markdown} & \textbf{HTML} \\
\midrule
GPT 5.2 & 91.47{\scriptsize±0.42} & 91.52{\scriptsize±0.62} & 91.54{\scriptsize±0.52} & \textbf{90.16}{\scriptsize±0.43} & 91.57{\scriptsize±0.63} \\
GPT 4.1 & 91.88{\scriptsize±1.35} & 91.11{\scriptsize±0.45} & 91.50{\scriptsize±0.65} & \textbf{89.03}{\scriptsize±0.40} & 91.53{\scriptsize±0.48} \\
Ministral 3 8B & 92.03{\scriptsize±0.28} & \textbf{58.03}{\scriptsize±0.95} & 90.60{\scriptsize±1.06} & 96.30{\scriptsize±0.40} & 88.37{\scriptsize±0.39} \\
GPT OSS 120B & 89.64{\scriptsize±0.49} & 89.62{\scriptsize±0.14} & 89.33{\scriptsize±0.74} & 88.50{\scriptsize±1.59} & \textbf{88.45}{\scriptsize±1.02} \\
Mistral Small 3.2 24B & 85.56{\scriptsize±0.25} & 87.16{\scriptsize±0.43} & \textbf{35.00}{\scriptsize±22.36} & 86.36{\scriptsize±0.45} & 82.16{\scriptsize±0.30} \\
GPT OSS 20B & 88.55{\scriptsize±0.84} & \textbf{80.48}{\scriptsize±17.30} & 88.48{\scriptsize±0.77} & 82.80{\scriptsize±9.39} & 86.85{\scriptsize±2.51} \\
Gemma 3 12B & \textbf{73.74}{\scriptsize±5.70} & 79.41{\scriptsize±5.97} & N/A & 74.20{\scriptsize±9.22} & 75.16{\scriptsize±6.69} \\
Gemma 3 4B & 75.57{\scriptsize±0.72} & 79.14{\scriptsize±0.79} & \textbf{60.57}{\scriptsize±2.75} & 73.20{\scriptsize±0.41} & 73.77{\scriptsize±0.80} \\
Llama 3.1 8B & \textbf{72.12}{\scriptsize±0.46} & 77.02{\scriptsize±0.84} & 73.56{\scriptsize±1.55} & 76.36{\scriptsize±0.68} & 73.33{\scriptsize±0.75} \\
\bottomrule
\end{tabular}
\caption{Overall Scores across format perturbation scores (mean ± std). N/A indicates complete failure to produce valid output. Note the high variance for Mistral Small 24B on XML (±22.36), indicating inconsistent behavior.\textbf{Bold} in data rows indicates lowest robustness for that model.}
\label{tab:format_scores}
\end{table*}

\textbf{Positional Perturbations.} Figure~\ref{fig:positional_scores} reveals dramatic differences in sensitivity to instruction ordering. The GPT 5 family shows stability (83.04.87--91.22) with minimal variance (±0.17--0.62). In contrast, GPT OSS models exhibit extremely high variance: ±18.93 for the 20B model and ±20.91 for the 120B model, indicating fundamentally unstable positional handling despite large scale. Smaller open-source models consistently struggle (Llama 3.1 8B: 63.33±0.52, Gemma 3 4B: 66.26±0.16).

\textbf{Format Perturbations.} Table~\ref{tab:format_scores} reveals format-specific failure modes invisible in aggregate scores. XML proves catastrophically difficult: Gemma 3 12B fails entirely to produce valid XML (N/A), Mistral Small 24B achieves only 35.00 with extreme variance (±22.36), and Gemma 3 4B scores 60.57±2.75. Notably, Ministral 3 8B—despite ranking third overall—fails on YAML (58.03±0.95). Top models (GPT 5.2, GPT 4.1) maintain uniform performance across all formats with low variance.

\textbf{Multilingual Perturbations.} Table~\ref{tab:multilingual} shows Quality Delta by language (lower is better). Japanese and Dutch emerge as especially challenging languages (mean delta: 44.32 and 44.31 respectively), followed by German (43.87). In particular, 6 out of the 11 models evaluated exhibit the largest quality delta on Japanese. Ministral 3 8B achieves remarkably uniform cross-language performance with the lowest deltas (26.15-31.31, range of only 5.16 points), while GPT 5.2 shows the highest quality deltas despite strong overall performance, suggesting a quality-consistency tradeoff.

\subsection{Universal vs. Model-Specific Difficulty}

Table~\ref{tab:category_difficulty} analyzes which perturbation types are universally challenging versus model-dependent.

\textbf{Universally hard:} Positional (-17.99), multilingual (-15.14), and crosslingual (-12.26) appear to be the most challenging perturbations across models. 

\textbf{Model-dependent:} Format perturbations show moderate mean difficulty but extreme variance on specific formats (XML, YAML). A model's format robustness cannot be predicted from overall performance—Ministral 3 8B's YAML failure demonstrates this clearly.

\textbf{Universally manageable:} General perturbations cause the smallest drops (-8.81) indicating predictable, manageable difficulty across all models tested.

\section{Conclusion}
We introduced a comprehensive benchmark evaluating LLM robustness across 11 models, four enterprise tasks, and five perturbation categories. Our evaluation of models ranging from 4B to 120B+ parameters yields several key findings:

\textbf{Perturbation hierarchy.} Models are least robust to multilingual and positional perturbations (15--17 point drops from baseline), moderately affected by format and crosslingual changes (10--12 points), and most resilient to general perturbations (9 points). This hierarchy is consistent across model sizes.

\textbf{Size-robustness relationship.} Conventional wisdom that ``bigger is better'' receives partial support but warrants qualification. Within model families, scaling tends to improve robustness (GPT 5: +6.66 points from nano to 5.2). However, training methodology appears to play a substantial role: Ministral 3 8B outperforms the 120B GPT OSS model, while Llama 3.1 8B—at identical 8B scale—performs worst overall. The 19.88-point gap between these 8B models demonstrates that architecture and training quality matter more than raw parameters.

\textbf{Format-specific vulnerabilities.} Even high-performing models exhibit catastrophic failures on specific formats. XML proves problematic for multiple models regardless of size, and Ministral 3 8B—despite ranking third overall—fails on YAML generation. Format robustness cannot be predicted from overall performance.

\textbf{Language challenges.} Japanese consistently emerges as the most difficult language across models. Crosslingual tasks (mismatched input/output languages) prove particularly challenging, with English-to-other-language generation showing unexpected weaknesses.

These findings offer actionable guidance for enterprise deployment: (1) robustness testing should be incorporated into model selection, particularly for multilingual and structured-output applications; (2) model size alone is an insufficient predictor of robustness—evaluation on specific perturbation types is necessary; (3) format requirements should be tested individually, as aggregate performance masks format-specific failures.


\section{Limitations}
While our benchmark provides a comprehensive evaluation of LLM robustness in enterprise scenarios, several limitations should be acknowledged:

\textbf{LLM-as-a-Judge:}
Our primary quality evaluation metrics for three of the four use cases rely on using GPT 4.1 to assess model responses. NLP experts tuned these judge prompts to ensure maximal consistency between human expert evaluation and LLM evaluation for the given metrics. While this approach enables scalable and consistent evaluation across a large dataset, it may overlook subtle errors that a human expert would detect. The LLM-as-a-judge paradigm, though increasingly common, is not a perfect substitute for human annotation, especially for nuanced or domain-specific tasks.

\textbf{Information Loss in Aggregation:}
To summarize robustness across tasks, models, and perturbation types, we aggregate results using overall scores and deltas. This aggregation can obscure important granular differences, such as specific failure modes or language-specific weaknesses, and may mask outlier behaviors that are critical in certain enterprise contexts. This consideration is particularly important when analyzing quality deltas and differences, as a net change of -6\% at the aggregate level does not clarify whether most individual records experienced minor quality declines or whether a mix of substantial decreases and offsetting increases contributed to the aggregate value.

\textbf{Non-Determinism in Model Outputs:}
Despite setting fixed random seeds and using temperature 0 for most models, we observed non-deterministic outputs across repeated runs. For some models (e.g., OpenAI reasoning models), temperature cannot be configured which may increase output variability. As a result, our robustness measurements may underestimate the true variance in model behavior, and some observed differences could be due to inherent sampling noise rather than prompt perturbations alone.

\textbf{Translation Quality and Verification:}
For multilingual perturbations, only the base prompt translations were reviewed and verified by linguistic experts familiar with each language being tested. The contextual data (e.g., case details, chat transcripts) was translated automatically by LLMs without human verification. This introduces the possibility of translation errors or artifacts that could affect model performance independently of the intended perturbation.

\textbf{Limited Close-Source Model Evaluations:} Due to budget constraints and the size of our experiments, we were only able to evaluate our robustness benchmarks on a small group of closed-source models from one company. Future work would ideally evaluate more frontier models from a wider range of labs to provide a more comprehensive view of robustness across the current LLM ecosystem. 

\textbf{Single Domain Focus:} The vast majority of the 420 base issues in our benchmark are IT-related software failures. This narrow domain focus may not capture robustness challenges present in other enterprise sectors such as healthcare, finance, legal, or manufacturing. Different domains may exhibit unique types of noise, jargon, and perturbation patterns that could reveal different robustness characteristics in LLMs.

\textbf{Limited Task Diversity:} The benchmark focuses on only 4 enterprise tasks: case summarization, chat summarization, question answering, and entity slot-filling. While these tasks are representative of common enterprise applications, they do not cover the full spectrum of LLM use cases in production environments, such as code generation, complex multi-step reasoning chains, data analysis, or creative content generation. 

\textbf{Isolated Perturbation Testing:} Each perturbation type is evaluated independently to isolate its specific effect on model behavior. However, real-world scenarios often involve multiple simultaneous perturbations—for example, a user might submit a misspelled query in a non-English language with unusual formatting. The compounding effects of multiple perturbations may reveal different robustness patterns than those observed when perturbations are applied in isolation.

\textbf{Limited Run Count:} Despite setting fixed random seeds and using temperature 0 for most models, we observed non-deterministic outputs across repeated runs and therefore conducted 5 independent runs per configuration. However, this may be insufficient to fully capture the variance in non-deterministic behaviors, particularly for models exhibiting high variance such as the GPT OSS family. A larger number of runs could provide more stable estimates of robustness, especially for models with inherent output variability.

\textbf{Fixed Temperature Settings:} All models were evaluated at temperature 0 (where supported) to maximize determinism and isolate the effects of perturbations from sampling variance. However, this does not reflect all real-world deployment scenarios where higher temperatures are often used to generate more diverse or creative outputs. We chose temperature 0 because higher temperatures would introduce additional output variance that could confound our robustness measurements—making it difficult to distinguish between perturbation-induced changes and temperature-induced stochasticity. Nevertheless, enterprises deploying models with higher temperature settings may observe different robustness characteristics than those reported here.

\textbf{Reasoning Model Limitations:} For reasoning-capable models, we used the lowest reasoning setting available to maintain consistency and control costs. This decision was informed by the relatively straightforward nature of our enterprise tasks, which do not require extensive chain-of-thought reasoning or complex problem decomposition. However, this configuration may not reflect optimal or typical deployment settings in production environments where higher reasoning settings might be employed for more complex tasks. Enterprises using higher reasoning settings may experience different robustness-performance tradeoffs than those captured in our evaluation.

\textbf{Language Coverage:} Our multilingual and cross-lingual evaluations include 8 languages: Spanish, French, French Canadian, German, Dutch, Italian, Japanese, and Brazilian Portuguese. While this provides meaningful coverage of major European and Asian languages, it excludes other widely-used languages in global enterprises such as Chinese, Hindi, Arabic, Korean, and various African and Southeast Asian languages. The robustness patterns we observe may not generalize to these underrepresented languages, which may have different linguistic characteristics, resource availability, and model training exposure.

\bibliography{custom}

\appendix
\section{Appendix}

\subsection{Model Versions}

We use the following versions of closed-source models in our experiments:

\begin{table}[h]
\centering
\begin{tabular}{ll}
\toprule
\textbf{Model} & \textbf{Version Used} \\
\midrule
GPT 5.2 & \texttt{2025-12-11} \\
GPT 5 Mini & \texttt{2025-08-07} \\
GPT 5 Nano & \texttt{2025-08-07} \\
GPT 4.1 & \texttt{2025-04-14} \\

\bottomrule
\end{tabular}
\caption{Closed-source model versions used in evaluation.}
\label{tab:model_versions}
\end{table}

\subsection{Similarity Metrics}
\label{appendix:similarity_metrics}

In addition to using an LLM as a judge to assess similarity between perturbed and unperturbed prompts, we also experimented with computing the following supplementary metrics. While these are not used in the core analysis presented in this paper, they are available for reference when running the benchmark. We found that these methods did not capture variation in core meaning and semantic relevance as well as LLM-as-a-Judge.

(1) \textbf{ROUGE-L} computes the longest common subsequence (LCS) between the original and perturbed responses. This metric reflects lexical similarity by capturing both precision and recall over sequence alignment, offering a surface-level comparison useful for detecting changes in phrasing or structure.

(2) \textbf{Semantic similarity}, measured using the \texttt{all-MiniLM-L6-v2} sentence transformer model. Each response is embedded into a dense vector space, and cosine similarity is calculated to quantify semantic overlap. This approach enables detection of meaning-preserving paraphrases and subtle shifts in content, even when lexical choices differ.

Note that most of the pormpts for each enterprise task expects a JSON-formatted output. To compute similarity in these cases, we first align the keys between the original and perturbed outputs, then calculate similarity scores on the corresponding values. We refer to these methods as \textbf{JSONRouge} and \textbf{JSONSimilarity}.

\subsection{AI Usage Disclosure}
We utilized AI assistants (Claude and ChatGPT) to support various aspects of this paper's preparation, including paraphrasing and refining sections for clarity, generating LaTeX formatting for tables and figures, proofreading and spell-checking, and conducting code review.

\subsection{Prompts}
\label{appendix:prompts}

\begin{tcolorbox}[title=Case Summarization Task Prompt, colframe=black, colback=white, ]
\textbf{User:}\\
You are given the following case that needs to be summarized: \\
\texttt{{\{\{case\_details\}\}}}

Be sure to include a few sections to make it easy for the user to scan. \\

Include the following sections:\\
Main Problem\\
- It should be the main issue that the case is about.\\
Key Events\\
- It should be a list of all the key events that took place over the course of the case\\
Final Resolution \\
- Any conclusion notes about the case that led to its resolution. \\

Only generate content for the above sections. \\
Output in the following JSON format with keys in English and values in English: \\
\texttt{\{
"Main Problem": <string>,
"Key Events": <list[string]>,
"Final Resolution": <string>\}}
\end{tcolorbox}

\begin{tcolorbox}[title=Entity Slot-Filling Task Prompt, colframe=black, colback=white, fontupper=\small]
\textbf{User:}

Here is a JSON schema of entities that may be able to be extracted from the case. \\

\texttt{\{\{entities\}\}} \\

Here is the text where entities may be extracted from : \\
      
Issue: \texttt{\{\{issue\}\}}

Question: \texttt{\{\{question\}\}} \\

Your task is to extract all of the entities that can be filled from the above entity schema from the text. \\

Please provide your output in the following JSON format: \\

\begin{verbatim}
{
  "entities": [
    {
      "name": <entity_name>,
      "value": <entity_value>
    }
  ]
}
\end{verbatim}

\end{tcolorbox}

\begin{tcolorbox}[title=Chat Summarization Task Prompt, colframe=black, colback=white, fontupper=\small]
\textbf{User:}\\
You are given the following chat transcript that needs to be summarized: \\
\texttt{{\{\{chat\_transcript\}\}}}

Be sure to include a few sections to make it easy for the user to scan. \\

Include the following sections:\\
Main Problem \\
- It should be the main issue that the case is about. \\
Key Events \\
- It should be a list of all the key events that took place over the course of the case \\
Final Resolution \\
- Any conclusion notes about the case that led to its resolution. \\
People Involved \\
- All the people and names of individuals who are involved in the chat \\
Sentiment \\
- The overall sentiment of the person interacting with the agent \\
Overall Summary \\
- A final 2-3 sentence summary to explain what went on during the chat \\

Only generate content for the above sections. \\
Output in the following JSON format with keys in English and values in English: \\

\texttt{\{
  "Main Problem": <string>,
  "Key Events": <list[string]>,\
  "Final Resolution": <string>,
  "People Involved": []\}}
\end{tcolorbox}


\begin{tcolorbox}[title=Question Answering Task Prompt, colframe=black, colback=white, fontupper=\small]
\textbf{User:}

You are given the following documents where an answer may be found.

\textbf{Doc 1:} \\
\texttt{\{\{kb\_document\_1\}\}}

\textbf{Doc 2:} \\
\texttt{\{\{kb\_document\_2\}\}}

\textbf{Doc 3:} \\
\texttt{\{\{kb\_document\_3\}\}}

\textbf{Doc 4:} \\
\texttt{\{\{kb\_document\_4\}\}}

The user has asked the following question: \\
\texttt{\{\{query\}\}}

Your job is to answer the user's question using \textbf{ONLY} the documents that are available to you.\\
Do not answer any questions that are not relevant to the above context.\\
This means that even if you did know the answer from your prior knowledge, you should not answer the question.

\textbf{Response Formatting}

Make sure that each sentence is cited appropriately using the following format (id).

For example: \\
Banks make money (doc\_1). Using that money, employees can purchase ice cream from the cafeteria. The cafeteria is at building B (doc\_2).

In the above case, \texttt{doc\_1} is where the "banks make money" statement comes from, while \texttt{doc\_2} contains information about ice cream and the cafeteria's location.

\textbf{NEVER} say a sentence that cannot be grounded in one of the above documents.

\textbf{Output Structure}

Provide your output in the following format with JSON keys in English and values in English: \\

\texttt{\{
  "output\_response": <string with inline citations>,\
  "references": <list[string]>
\}}

\end{tcolorbox}

\begin{tcolorbox}[title=Robustness Judge Prompt, colframe=black, colback=white, fontupper=\small]
\textbf{System:}\\
You are a expert focused on evaluating how closely a candidate response aligns with a reference response.\\
\textbf{User:}\\
Please consider the following factors in your evaluation:\\
Does the content presented in the candidate response match the content of the reference response? \\
Are there any significant differences in the details and entities presented in each response? \\
Details and entities refer to any proper nouns and specific pieces of information. \\ 

\texttt{<BEGIN\_REFERENCE\_RESPONSE>}

\texttt{{\{\{reference\_response\}\}}}

\texttt{<END\_REFERENCE\_RESPONSE>} \\

\texttt{<BEGIN\_CANDIDATE\_RESPONSE>}

\texttt{{\{\{candidate\_response\}\}}}

\texttt{<END\_CANDIDATE\_RESPONSE>}\\

\textbf{Rating Criteria:} \\
3 - Strongly Aligned: The candidate response contains the same content as the reference response. Both responses include the same details and entities. Paraphrasing is fine as long as the details and entities are the same.\\
2 - Moderately Aligned: The candidate responses conveys the same ideas as the reference response, but there are some differences in the specific details and entities included in each response. \\
1 - Not Aligned: The candidate response conveys conveys a different message than the reference response. \\
The details and entities in the candidate response do not match the details and entities in the reference response. \\
\textbf{OUTPUT FORMAT:} \\
  Please provide the response in a valid JSON format as shown below: \\
  \texttt{\{
      "rating": <your score here>,
      "explanation": "<explanation of how closely the responses align, mention any notable discrepancies here>"\}}
\end{tcolorbox}

\begin{tcolorbox}[title=Completeness Judge Prompt (Case Summarization), colframe=black, colback=white, fontupper=\small]
\textbf{User:}\\
You are an expert judge tasked with judging the completeness of an AI generated summary of case details. \\
To make your assessment, the following will be provided below.
Evaluation criteria, Rating criteria, AI generated summary, Case Details, Output Format

Follow the steps below to judge the completeness of the AI generated summary. \\

\textbf{AI GENERATED SUMMARY}: \\

\texttt{{\{\{response\}\}}} \\

First, identify the sections in the AI generated summary above. \\

\textbf{CASE DETAILS}: \\

\texttt{{\{\{case\_details\}\}}} \\

Second, compare the AI generated summary to the case details.

For each section of the summary, consider whether the content of the section captures the relevant details and entities from the case details.

Entities refer to things like case numbers, contact information, account numbers, and other case specific information. 

Key entities are the entities that are most critical for understanding the case. \\

Now you will rate the completeness of the AI generated summary using the EVALUATION CRITERIA and RATING CRITERIA below. \\

\textbf{EVALUATION CRITERIA}:
\begin{itemize}
\item The summary includes key details and the most relevant entities from the case details
\item The summary does NOT need to include every detail from the case details
\item For each section of the summary, that section content includes the key entities for the details it describes.
\end{itemize}

\textbf{RATING CRITERIA}: \\
Rate the completeness of the generated summary on a scale of 1-3. \\
1 - the summary sections contain very few or none of the relevant details and entities from the case details.

2 - the summary sections contains some relevant details and entities from the case details but miss the most relevant details and entities for some sections of summary.

3 - all sections of the summary include the most relevant details and entities. The summary contains many key details and entities from the case details.

\textbf{OUTPUT FORMAT:} \\
  Please provide the response in a valid JSON format as shown below: \\
  \texttt{\{
      "explanation": <your\_explanation\_for\_the\_rating\_here>,
      "rating": "<your\_score\_here>"\}}
\end{tcolorbox}

\begin{tcolorbox}[title=Faithfulness Judge Prompt (Chat Summarization), colframe=black, colback=white, fontupper=\small]
\textbf{User:}\\
You are an expert judge tasked with judging the faithfulness of an AI generated summary of a chat transcript. \\
To make your assessment, the following will be provided below. \\
\begin{itemize}
\item  Evaluation criteria
\item  Rating criteria
\item  AI generated summary
\item Chat Transcript
\item Output Format
\end{itemize}

Use the information below to judge the faithfulness of the AI generated summary. \\

\textbf{EVALUATION CRITERIA}:
\begin{itemize}
\item The summary only includes details and information provided in the chat transcript
\item The summary only includes assumptions that can be reasonably implied from the chat transcript
\item The summary DOES NOT misinterpret or misrepresent information from the chat transcript
\item Failure to include relevant details from the chat transcript DOES NOT affect faithfulness
\end{itemize}

\textbf{RATING CRITERIA}: \\
Rate the faithfulness of the generated summary on a scale of 1-3. Remember that failure to include relevant details should NOT affect the rating and should NOT be mentioned in the explanation. \\

1 - the summary contains major hallucinations, includes details not mentioned in the chat transcript, or is completely unrelated to the chat transcript.

2 - the summary contains minor hallucinations such as misinterpreting or misrepresenting details from the chat transcript.

3 - the summary contains no hallucinations and the summary is grounded in the chat transcript. All interpretations made by the summary are have some basis in the chat transcript and are logical assumptions. \\
      
\textbf{AI GENERATED SUMMARY}: \\

\texttt{{\{\{response\}\}}} \\

\textbf{CHAT TRANSCRIPT}: \\

\texttt{{\{\{chat\_transcript\}\}}} \\

\textbf{OUTPUT FORMAT:} \\
  Please provide the response in a valid JSON format as shown below: \\
  \texttt{\{
      "explanation": <your\_explanation\_for\_the\_rating\_here>,
      "rating": "<your\_score\_here>"\}}
\end{tcolorbox}

\begin{tcolorbox}[title=Relevance Judge Prompt (Question Answering), colframe=black, colback=white, fontupper=\small]
\textbf{User:}\\
You have the following documents which are related to the user query: doc\_1, doc\_2, doc\_3 and etc. \\

\textbf{\*\***DOCUMENTS\***} \\
\textbf{ Doc 1 } \\

\texttt{{\{\{kb\_1\}\}}} \\

\textbf{ Doc 2 }\\

\texttt{{\{\{kb\_2\}\}}} \\

\textbf{ Doc 3 } \\

\texttt{{\{\{kb\_3\}\}}} \\

\textbf{ Doc 4 }\\

\texttt{{\{\{kb\_4\}\}}} \\

\textbf{\***END\_DOCUMENTS\***} \\

You are also provided with the below user query: \\

\textbf{\***QUERY\***} \\

\texttt{{\{\{query\}\}}} \\

\textbf{\***END\_QUERY\***} \\

Your task is to grade the response generated by the agent to the user query. \\

The response is provided below: \\

\textbf{\***AGENT\_RESPONSE\***} \\

\texttt{{\{\{response\}\}}} \\

\textbf{\***END\_AGENT\_RESPONSE\***} \\

\end{tcolorbox}

\begin{tcolorbox}[title=Relevance Judge Prompt (Question Answering) cont'd, colframe=black, colback=white, fontupper=\small]

\***\textbf{Your Task}\***
      
Evaluate the relevance of the response to the user query. 

Relevant means closely connected or appropriate to what is being done or considered. Information may be incorrect or inaccurate, that is not related to relevance.

To do this task, start by meticulously analyzing the agent's response against the user query and documents. The analysis should be comprehensive and cover all the information in the response that is relevant to the user query. Then proceed by assigning one of the scores below:

- 3: In this case, all the information provided in the agent response, every single sentence needs to be analyzed, are relevant. Furthermore, at least 50\% of the response needs to be directly addressing the user query. This means that all the information in the response should be relevant in some sense but we allow for a maximum half of the response not to be directly relevant. Any evidence of relevance should be included in your analysis.

- 2: In this case, all the information provided in the agent response, every single sentence needs to be analyzed, are relevant. However, more than half of the response is not directly addressing the user's query.

- 1: In this case, the response contains information not relevant to the user query.

\textbf{EXAMPLE}
\begin{quote}
\textbf{USER\_QUERY}

\begin{verbatim}
user: Is the sky blue?
\end{verbatim}

\textbf{END\_USER\_QUERY}

\textbf{AGENT\_RESPONSE}

Yes, the sky often appears blue during the day.
This is due to a phenomenon called Rayleigh scattering. 
When sunlight enters Earth's atmosphere, it collides with molecules
and small particles in the air. 
Blue light is scattered in all directions by the gases and particles 
in the atmosphere more than other colors because it travels as shorter,
smaller waves. 
This scattered blue light is what we see when we look up at the sky.

\textbf{END\_AGENT\_RESPONSE}

\textbf{Analysis:} The response starts by directly responding to the user query. However, it proceeds by four sentence explaining why the sky is blue which is not directly addressing the query. Therefore, the entire response is relevant but only the first sentence is directly addressing the user's query. Therefore, based on the descriptions, it is acceptable. \\

\textbf{Rating:} Acceptable

\textbf{END\_EXAMPLE}
\end{quote}

\end{tcolorbox}

\begin{tcolorbox}[title=Relevance Judge Prompt (Question Answering) cont'd, colframe=black, colback=white, fontupper=\small]

\textbf{EXAMPLE}
\begin{quote}
\textbf{USER\_QUERY}

\begin{verbatim}
user: Is the sky blue?
\end{verbatim}

\textbf{END\_USER\_QUERY}

\textbf{AGENT\_RESPONSE}

Yes, the sky often appears blue during the day.

\textbf{END\_AGENT\_RESPONSE}

\textbf{Analysis:} The response directly answers the user query. It does not contain any extra information and therefore, is ideal according to the descriptions. \\

\textbf{Rating:} Ideal

\textbf{END\_EXAMPLE}
\end{quote}

\textbf{EXAMPLE}
\begin{quote}
\textbf{USER\_QUERY}

\begin{verbatim}
user: Is the sky blue?
\end{verbatim}

\textbf{END\_USER\_QUERY}

\textbf{AGENT\_RESPONSE}

Yes, the sky often appears blue during the day.
This is due to a phenomenon called Rayleigh scattering. 
When sunlight enters Earth's atmosphere, it collides with
molecules and small particles in the air.
Blue light is scattered in all directions by the gases 
and particles in the atmosphere more than other colors because
it travels as shorter, smaller waves. 
This scattered blue light is what we see when we look up at the sky. 
We should note that blue is the favourite colour among
most successful men and represents calmness and power inspired by the sky.

\textbf{END\_AGENT\_RESPONSE}

\textbf{Analysis:} The response starts by directly responding to the user query. However, it proceeds by four sentence explaining why the sky is blue which is not directly addressing the query. Furthermore, it contains an irrelevant sentence at the end. Therefore, the response is unacceptable because of the irrelevant sentence. \\

\textbf{Rating:} Unacceptable

\textbf{END\_EXAMPLE}
\end{quote}

\textbf{OUTPUT FORMAT:} \\
  Please provide the response in a valid JSON format as shown below: \\
  \texttt{\{
      "explanation": <your\_explanation\_for\_the\_rating\_here>,
      "rating": "<your\_score\_here>"\}}
\end{tcolorbox}

\begin{tcolorbox}[title=Conciseness Judge Prompt (Question Answering), colframe=black, colback=white, fontupper=\small]
\textbf{User:}\\
You are provided following documents served as the knowledge base for the user's question: \\

\textbf{\*\***DOCUMENTS\***} \\
\textbf{ Doc 1 } \\

\texttt{{\{\{kb\_1\}\}}} \\

\textbf{ Doc 2 } \\

\texttt{{\{\{kb\_2\}\}}} \\

\textbf{ Doc 3 }\\

\texttt{{\{\{kb\_3\}\}}} \\

\textbf{ Doc 4 }\\

\texttt{{\{\{kb\_4\}\}}} \\

\textbf{\***END\_DOCUMENTS\***} \\

You are also provided with the below user query: \\

\textbf{\***QUERY\***} \\

\texttt{{\{\{query\}\}}} \\

\textbf{\***END\_QUERY\***} \\

And you have the following response from an agent to answer the user's query. \\

\textbf{\***AGENT\_RESPONSE\***} \\

\texttt{{\{\{response\}\}}} \\

\textbf{\***END\_AGENT\_RESPONSE\***} \\

\***\textbf{Your Task}\***
      
Your task is to evaluate if the agent's response is sufficiently concise. \\
      
- For more specific queries: The response should include the necessary details from the retrieved context to fulfill the user's intent or answer the question. \\
- For generic queries: The response should contain the topically relevant details from the retrieved context. Example of generic queries "401k", "benefits", "access" etc \\

\end{tcolorbox}

\begin{tcolorbox}[title=Conciseness Judge Prompt (Question Answering) cont'd, colframe=black, colback=white, fontupper=\small]

Review the response carefully and consider these factors (in the order of importance): \\

1. Redundancy: Is there repetitive information or extra (non-informative or empathetic) language?

2. Answer length: General questions should be answered with a general overview with no specific details. Specific queries should not be too long, but can be longer than those of the generic queries and are more detailed.

3. Specificity: Does the level of detail match the question - do specific queries have specific answers and do generic queries have more generic answers? Are there overly specific details for a general question?

4. Response to generic queries: When a query is ambiguous, each bullet point in the response should not describe what the document is about, but rather the information contained therein. If the query is specific, rate this factor as "good"

For each factor, rate it as: \\

- Good: No issues found

- Fair: Minor issues that don't significantly impact concision

- Poor: Significant issues that make the response less concise \\

Based on your evaluation, assign one of the following scores: \\

- 3: All factors are rated as Good, or at most one is Fair

- 2: Most factors are Good or Fair, with at most one Poor

- 1: Two or more factors are rated as Poor \\

\textbf{OUTPUT FORMAT:} \\
  Please provide the response in a valid JSON format as shown below: \\
  \texttt{\{
      "explanation": <your\_explanation\_for\_the\_rating\_here>,
      "rating": "<your\_score\_here>"\}}

\end{tcolorbox}

\begin{tcolorbox}[title=Citation Judge Prompt (Question Answering), colframe=black, colback=white, fontupper=\small]
\textbf{User:}\\
You are tasked with judging whether a response contains appropriate citations given the reference documents.

\textbf{\*\***REFERENCE\_DOCUMENTS\***} \\

\textbf{ Doc 1 }
\texttt{{\{\{kb\_1\}\}}} \\

\textbf{ Doc 2 }
\texttt{{\{\{kb\_2\}\}}} \\

\textbf{ Doc 3 }
\texttt{{\{\{kb\_3\}\}}} \\

\textbf{Doc 4 }\texttt{{\{\{kb\_4\}\}}} \\

\textbf{\***END\_REFERENCE\_DOCUMENTS\***} \\

\textbf{\***RESPONSE\***} \\

\texttt{{\{\{response\}\}}} \\

\textbf{\***END\_RESPONSE\***} \\

\textbf{\***EVALUATION\_CRITERIA\***} \\

Consider each sentence in the RESPONSE. \\
If the sentence makes a factual claim, check whether it has a corresponding citation at the end. \\
If the sentence does contain a citation, refer to that document and check if the factual claim is supported by information in the corresponding reference document. \\
For each sentence, if it requires a citation and the citation is supported by the reference document, that sentence is considered acceptable. \\

If the sentence requires a citation but the citation is not supported by the reference document, that sentence is considered unacceptable. \\

If the sentence does not require a citation, that sentence is considered acceptable. \\

Once you have evaluated each sentence, determine the overall score for the response given the guidelines below: \\

- 3: All sentences are rated as acceptable

- 2: The majority of the sentences are rated as acceptable.

- 1: Less than half of the sentences are rated as acceptable \\

\textbf{\***END\_EVALUATION\_CRITERIA\***} \\

\textbf{OUTPUT FORMAT:} \\
  Please provide the response in a valid JSON format as shown below: \\
  \texttt{\{
      "explanation": <your\_explanation\_for\_the\_rating\_here>,
      "rating": "<your\_score\_here>"\}}

\end{tcolorbox}

\subsection{Dataset Samples}
We show below samples from our dataset for each use case.

\begin{table*}[t]
\centering
\begin{tabular}{|l|p{16cm}|l|l|}
\hline
\textbf{ID} & \textbf{Case Details} \\
\hline
160 &  
Terraform apply fails due to missing state lock release.  
Multiple apply attempts blocked with error "Resource is locked by another process".  
\textit{Root cause:} Dangling lock after interrupted process.  
\textit{Resolution:} Used `terraform force-unlock`, released lock, deployments unblocked.   \\
\hline
321 &  
ECR image scan not updating after new image push.  
Customer's new images pushed to Amazon ECR show "Scan status: Not scanned" despite scan-on-push enabled and manual scans. Found to be caused by the lifecycle policy deleting images before scan finishes.  
\textit{Resolution:} Increased retention period in policy; scan completes and status updates as expected.   \\
\hline
369 &  
OAuth authentication failing due to redirect URI mismatch after promoting to production.  
Production logins failed with "redirect\_uri\_mismatch", worked in staging.  
\textit{Root cause:} OAuth config for production missing authorized redirect URI.  
\textit{Resolution:} Added correct URI; logins now work in both environments.  \\
\hline
\end{tabular}
\caption{Case Summarization - Sample Case Details. Case details are truncated for readability.}
\label{tab:case_details}
\end{table*}

\begin{table*}[htbp]
\centering
\begin{tabular}{|l|p{16cm}|}
\hline
\textbf{ID} & \textbf{Chat Transcript} \\
\hline
160 &  
[15:02] Virtual Agent: Hello! I am here to help you with your Terraform issues. \newline
[15:03] Jeremy Clark: Hi, my terraform apply keeps failing. It says something about a state lock not being released. \newline
[15:04] Virtual Agent: It sounds like the state lock may not have released properly due to the interruption. Would you like to try unlocking the state file automatically? \newline
[15:05] Jeremy Clark: Yes, please. \newline
[15:06] System: Nicole Daniels has joined the chat. \newline
[15:06] Nicole Daniels: Hi Jeremy, I see you’re running into an S3 state lock issue. Could you please share the exact error message you’re seeing? \newline
[15:06] Jeremy Clark: Sure. It says: Error acquiring the state lock: Error message: state file is locked by another process. \\
\hline
321 &  
[11:35] Virtual Agent: Hello! Welcome to AWS Support Chat. \newline
[11:36] Sameer Patel: Hi, I pushed a new image to my ECR repo, but the image scan status isn't updating. It still shows the previous scan result. \newline
[11:37] Virtual Agent: Are image scanning settings enabled for "api-backend"? \newline
[11:37] Sameer Patel: Yes, scanOnPush is enabled for this repo. \newline
[11:38] Virtual Agent: Got it. Let me connect you with a live agent for further investigation. \newline
[11:38] Lauren Choi: Hi Sameer, this is Lauren from AWS Support. I see you're having trouble with ECR image scan results not updating after a new image push. \\
\hline
369 &  
[09:15] Virtual Agent: Hello! How can I assist you today? \newline
[09:16] Marissa Gilbert: Hi, I'm running into an OAuth redirect URI mismatch error in our staging environment, but prod is working fine. \newline
[09:17] Virtual Agent: Understood. Are you using the same OAuth provider for both environments? \newline
[09:18] Marissa Gilbert: Yes, Google OAuth for both, but with separate client IDs for staging and production. \newline
[09:19] Marissa Gilbert: I just checked Google Cloud Console. For staging I have "https://staging.ourapp.com/oauth/callback" registered. The app is using that exact URL. \newline
[09:29] Marissa Gilbert: Thanks for your help! \newline
[09:29] Virtual Agent: You’re welcome! Have a great day. \\
\hline
\end{tabular}
\caption{Chat Summarization - Sample Chat Transcripts. Utterances truncated readability.}
\label{tab:chat_transcript}
\end{table*}

\begin{table*}[t]
\centering
\begin{tabular}{|l|p{5.5cm}|p{9.5cm}|}
\hline
\textbf{ID} & \textbf{Question} & \textbf{Knowledge Base Documents} \\
\hline
160 &  
I'm encountering Terraform state lock errors, and I'm not sure if it's due to my network being unreliable or because multiple operations are interfering with each other. How can I effectively diagnose and resolve these issues? 
& 
KB 1 Title: Troubleshooting Terraform Apply Failure Due to State Lock Not Released.

KB 2 Title: Understanding and Preventing Terraform State Lock Contention Errors.

KB 3 Title: Diagnosing and Resolving Orphaned Terraform State Locks in Remote Backends. \\
\hline

321 & 
I'm noticing that after pushing a new image to ECR, my scan results don't update or still show outdated vulnerabilities. What could be causing this issue even though my repository has scan-on-push enabled?
&
KB 1 Title : ECR Image Scan Not Updating After New Push.

KB 2 Title : Resolving ECR Image Scan Results Showing Old Vulnerabilities After New Image Push.

KB 3 Title : Understanding ECR Image Scan Behavior with Immutable Tags and Multi-Architecture Images
\\
\hline

369 & 
I'm encountering intermittent 'redirect\_uri\_mismatch' errors during deployments across my staging and production environments, and the issue seems to occur inconsistently for different users. Can you explain the common causes and suggest measures to both resolve and prevent these errors?
&
KB 1 Title : Resolving OAuth Redirect URI Mismatch Between Staging and Production.

KB 2 Title : Preventing OAuth Redirect URI Issues When Switching Between Staging and Production Environments.

KB 3 Title : Understanding Security Implications of OAuth Redirect URI Mismatch Across Staging and Production.
\\
\hline

\end{tabular}
\caption{Question Answering - Sample Questions and Knowledge Base Document Titles. KB Document content removed for readability}
\label{tab:qna_kb_trunc}
\end{table*}

\begin{table*}[t]
\centering
\begin{tabular}{|c|p{3.3cm}|p{6.2cm}|p{4.5cm}|}
\hline
\textbf{ID} & \textbf{Question} & \textbf{Entity Schema (name, type: description)} & \textbf{Non-null Extracted Entities} \\
\hline
160 &
I tried running a Terraform apply operation, but it keeps failing because of a missing state lock release. I'm using Terraform version 1.0.11 with a remote backend. It's happening in our staging environment. Can you help me understand why this is occurring?
&
\textbf{error\_message} (\textit{string}: The error message received during the operation)\newline
\textbf{resource\_name} (\textit{string}: Name of the resource being managed)\newline
\textbf{operation\_type} (\textit{choice} [apply, plan, destroy]: Type of operation being performed)\newline
\textbf{state\_lock\_status} (\textit{choice} [locked, unlocked]: Status of the state lock)\newline
\textbf{terraform\_version} (\textit{string}: Version of Terraform being used)\newline
\textbf{backend\_type} (\textit{choice} [local, remote]: Type of backend used for storing state)\newline
\textbf{retry\_attempts} (\textit{number}: Number of retry attempts made)\newline
\textbf{timeout\_duration} (\textit{number}: Duration of timeout for the operation)\newline
\textbf{user\_permissions} (\textit{string}: Permissions of the user executing the operation)\newline
\textbf{environment} (\textit{choice} [development, staging, production]: Environment in which the operation is being performed)\newline
\textbf{issue\_details} (\textit{other}: Detailed description of the issue)
&
\textbf{error\_message}: missing state lock release\newline
\textbf{operation\_type}: apply\newline
\textbf{terraform\_version}: 1.0.11\newline
\textbf{backend\_type}: remote\newline
\textbf{environment}: staging\newline
\textbf{issue\_details}: apply fails due to missing state lock release
\\
\hline

321 &
Hi, I pushed a new Docker image to our ECR repository named "app-backend" last night, but the scan status is still showing as "pending." This is in the us-west-2 region. Could you help me understand why the scan isn't updating? My AWS account ID is 123456789012. Thanks!
&
\textbf{repository\_name} (\textit{string}: Name of the ECR repository)\newline
\textbf{image\_tag} (\textit{string}: Tag of the Docker image)\newline
\textbf{scan\_status} (\textit{choice} [pending, in-progress, complete, failed]: Status of image scan)\newline
\textbf{scan\_results} (\textit{other}: Results of the image scan)\newline
\textbf{push\_timestamp} (\textit{string}: Timestamp when the image was pushed)\newline
\textbf{scan\_timestamp} (\textit{string}: Timestamp when the scan was performed)\newline
\textbf{scan\_trigger} (\textit{choice} [manual, automatic]: Trigger for the image scan)\newline
\textbf{error\_message} (\textit{string}: Error message if scan failed)\newline
\textbf{aws\_region} (\textit{choice} [us-east-1, us-west-2, eu-west-1, ap-southeast-1]: Region of ECR repository)\newline
\textbf{account\_id} (\textit{string}: AWS account ID)
&
\textbf{repository\_name}: app-backend\newline
\textbf{scan\_status}: pending\newline
\textbf{aws\_region}: us-west-2\newline
\textbf{account\_id}: 123456789012
\\
\hline
\end{tabular}
\caption{Entity Slot-Filling - Sample questions, entity schema (with types and descriptions), and ground-truth extracted entities.}
\label{tab:entity_schema_ids}
\end{table*}

\end{document}